\documentclass{article} 
\usepackage{iclr2020_conference,times}

\usepackage{amsmath,amsfonts,bm}

\def\eqref#1{equation~\ref{#1}}

\def\1{\bm{1}}

\def\va{{\bm{a}}}

\def\vo{{\bm{o}}}

\def\vs{{\bm{s}}}

\DeclareMathAlphabet{\mathsfit}{\encodingdefault}{\sfdefault}{m}{sl}
\SetMathAlphabet{\mathsfit}{bold}{\encodingdefault}{\sfdefault}{bx}{n}

\newcommand{\E}{\mathbb{E}}

\newcommand{\KL}{D_{\mathrm{KL}}}

\usepackage{multirow}
\usepackage{array}
\usepackage{hyperref}
\usepackage{url}
\usepackage{rotating}
\usepackage{subcaption}
\usepackage{appendix}
\usepackage{todonotes}
\usepackage{amssymb}
\usepackage{amsmath}

\title{Deep Active Inference\\ for Autonomous Robot Navigation}

\author{Ozan \c{C}atal, Samuel Wauthier, Tim Verbelen, Cedric De Boom, \& Bart Dhoedt\\
IDLab, Department of Information Technology \\
Ghent University --imec\\
Ghent, Belgium \\
\texttt{ozan.catal@ugent.be}
}

\iclrfinalcopy 
\begin{document}

\maketitle

\begin{abstract}
Active inference is a theory that underpins the way biological agent's perceive and act in the real world. At its core, active inference is based on the principle that the brain is an approximate Bayesian inference engine, building an internal generative model to drive agents towards minimal surprise. Although this theory has shown interesting results with grounding in cognitive neuroscience, its application remains limited to simulations with small, predefined sensor and state spaces.

In this paper, we leverage recent advances in deep learning to build more complex generative models that can work without a predefined states space. State representations are learned end-to-end from real-world, high-dimensional sensory data such as camera frames. We also show that these generative models can be used to engage in active inference. To the best of our knowledge this is the first application of deep active inference for a real-world robot navigation task.
\end{abstract}

\section{Introduction}\label{intro}
Active inference and the free energy principle underpins the way our brain -- and natural agents in general -- work. The core idea is that the brain entertains a (generative) model of the world which allows it to learn cause and effect and to predict future sensory observations. It does so by constantly minimising its prediction error or ``surprise'', either by updating the generative model, or by inferring actions that will lead to less surprising states. As such, the brain acts as an approximate Bayesian inference engine, constantly striving for homeostasis.

There is ample evidence~\citep{FristonBioSys,FristonChoice,Friston2014ChoiceDope} that different regions of the brain actively engage in variational free energy minimisation. Theoretical grounds indicate that even the simplest of life forms act in a free energy minimising way~\citep{Friston2013life}.

Although there is a large body of work on active inference for artificial agents~\citep{Friston2006,FristonMountaincar,Friston2017,Friston2013,Cullen2018}, experiments are typically done in a simulated environment with predefined and simple state and sensor spaces. Recently, research has been done on using deep neural networks as an implementation of the active inference generative model, resulting in the umbrella term ``deep active inference''. However, so far all of these approaches were only tested on fairly simple, simulated environments~\citep{Ueltzhoffer2018,Millidge2019,Catal2019}. In this paper, we apply deep active inference on a robot navigation task, with high-dimensional camera observations and deploy it on a mobile robot platform. To the best of our knowledge, this is the first time that active inference is applied on a real-world robot navigation task.

In the remainder of this paper we will first introduce the active inference theory in Section~\ref{ai}. Next, we show how we implement active inference using deep neural networks in Section~\ref{nn}, and discuss initial experiments in Section~\ref{results}.

\section{Active Inference}\label{ai}
Active inference is a process theory of the brain that utilises the concept of free energy~\citep{Friston2013life} to describe the behaviour of various agents. It stipulates that all agents act in order to minimise their own uncertainty of the world. This uncertainty is expressed as Bayesian Surprise, or alternatively the variational free energy. In this context this is characterised by the difference between what an agent imagines about the world and what it has perceived about the world~\citep{Friston2010}.
More concretely, the agent builds a generative model $P(\tilde{\vo}, \tilde{\vs}, \tilde{\va})$, linking together the agents internal belief states $\vs$ with the perceived actions $\va$ and observations $\vo$ in the form of a joint distribution. We use a tilde to denote a sequence of variables through time. This generative model can be factorised as in Equation~\ref{eq:factor}.
\begin{equation}
   \label{eq:factor}
   P(\tilde{\vo}, \tilde{\vs}, \tilde{\va}) = P(\tilde{\va})P(\vs_0)\prod_{t=1}^{T}P(\vo_t|\vs_t)P(\vs_{t}|\vs_{t-1}, \va_{t-1})
\end{equation}

The free energy or Bayesian surprise is then defined as: 
\begin{equation}
   \label{eq:free-energy}
   \begin{split}
      F &= \E_Q [ \log Q(\tilde{\vs}) - \log P(\tilde{\vo}, \tilde{\vs}, \tilde{\va})] \\
        &= \KL (Q(\tilde{\vs}) \Vert P(\tilde{\vs}, \tilde{\va} | \tilde{\vo}) ) - \log P(\tilde{\vo}) \\
        &= \KL (Q(\tilde{\vs}) \Vert P(\tilde{\vs}, \tilde{\va})) - \E_{Q} [ \log P(\tilde{\vo} | \tilde{\vs})]
   \end{split}
\end{equation}

Here, $Q(\tilde{\vs})$ is an approximate posterior distribution. The second equality shows that free energy is equivalent to the (negative) evidence lower bound (ELBO)~\citep{Kingma13,Rezende14}. The final equation frames the problem of free energy minimisation as explaining the world from the agents beliefs whilst minimising the complexity of accurate explanations~\citep{Friston2016Learning}.

Crucially, in active inference agents will act according to the belief that they will keep minimising surprise in the future. 
This means agents will infer policies that yield minimal expected free energy in the future, with a policy $\pi$ being the sequence of future actions $\va_{t:t+H}$ starting at current time step $t$ with a time horizon $H$.
This principle is formalised in Equation~\ref{eq:G} with $\sigma$ being the softmax function with precision parameter $\gamma$.
\begin{equation}
   \label{eq:G}
   \begin{split}
      P(\pi) &= \sigma(-\gamma G (\pi)) \\
      G(\pi) &= \sum_{\tau=t}^{t+H} G(\pi, \tau)
   \end{split}
\end{equation}
Expanding the expected free energy functional $G(\pi, \tau)$ we get Equation~\ref{eq:G-complete}. Using the factorisation of the generative model from Equation~\ref{eq:factor} we approximate $Q(\vo_{\tau}, \vs_{\tau} | \pi) \approx P(\vo_{\tau} | \vs_{\tau})Q(\vs_{\tau} | \pi)$.
\begin{equation}
   \label{eq:G-complete}
   \begin{split}
      G(\pi, \tau) &= \E_{Q(\vo_{\tau}, \vs_{\tau} \vert \pi)} [ \log Q(\vs_{\tau} \vert \pi) - \log P(\vo_{\tau}, \vs_{\tau} \vert \pi)] \\
                   &= \E_{Q(\vo_{\tau}, \vs_{\tau} \vert \pi)} [ \log Q(\vs_{\tau} \vert \pi) - \log P(\vo_\tau \vert \vs_\tau, \pi) - \log P(\vs_\tau \vert \pi)] \\
                   &= \KL (Q(\vs_\tau \vert \pi) \Vert P(\vs_\tau)) + \E_{Q(\vs_\tau)} [ H(P(\vo_\tau \vert \vs_\tau))]
  \end{split}
\end{equation}

Note that, in the final equality, we substitute $P(\vs_{\tau} \vert \pi)$ by $P(\vs_{\tau})$, a global prior distribution on the so-called ``preferred'' states of the agent. This reflects the fact that the agent has prior expectations about the states it will reach. Hence, minimising expected free energy entails both realising preferences, while minimising the ambiguity of the visited states.

\section{Deep active inference}\label{nn}
\begin{figure}[t!]
   \centering
   \includegraphics[width=3in]{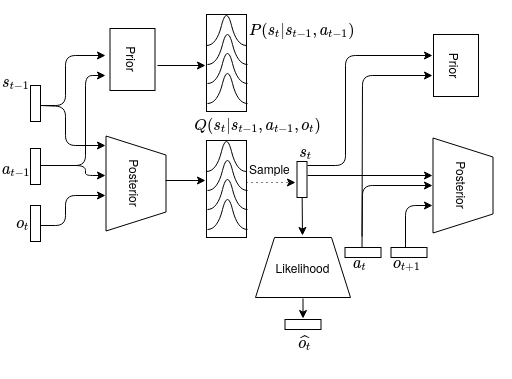}
   \caption{The various components of the agent rolled out trough time. We minimise the variational free energy by minimising both the negative log likelihood of observations and the KL divergence between the state transition model and the observation model. The inferred hidden state is characterised as a multivariate Gaussian distribution.}\label{fig:training}
\end{figure}

In current treatments of active inference the state spaces are typically completely fixed upfront~\citep{FristonMountaincar,Millidge2019} or partially~\citep{Ueltzhoffer2018}. However, this does not scale well for more complex tasks as it is often difficult to design meaningful state spaces for such problems. Therefore we allow for the agent to learn by itself what the exact parameterisation of its belief space should be. We enable this by using deep neural networks to generate the various necessary probability distributions for our agent.

We approximate the variational posterior distribution for a \emph{single} timestep  $Q(\vs_t \vert \vs_{t-1}, \va_{t-1}, \vo_{t})$ with a network $q_\phi(\vs_t \vert \vs_{t-1}, \va_{t-1}, \vo_{t})$. Similarly we approximate the likelihood model $P(\vo_t \vert \vs_t)$ with the network $p_\xi(\vo_t \vert \vs_t)$ and the prior $P(\vs_{t} \vert \vs_{t-1}, \va_{t-1})$ with the network $p_\theta(\vs_{t} \vert \vs_{t-1}, \va_{t-1})$. Each of the networks output a multivariate normal distribution with a diagonal covariance matrix using the reparameterisation trick~\citep{Kingma13}. These neural networks cooperate in a way similar to a VAE, where the fixed standard normal prior is replaced with the learnable prior $p_\theta$, the decoder by $p_\xi$ and finally the encoder by $q_\phi$, as visualised in Figure~\ref{fig:training}.

These networks are trained end-to-end using the free energy formula from the previous section as an objective.
\begin{equation}
   \label{eq:model-obj}
   \forall t : \underset{\phi, \theta, \xi}{\text{minimise}}: -\log p_\xi(\vo_t \vert \vs_t)
   + \KL (q_\phi(\vs_t \vert \vs_{t-1}, \va_{t-1}, \vo_{t}) \Vert p_\theta(\vs_{t} \vert \vs_{t-1}, \va_{t-1}))
\end{equation}
As in a conventional VAE~\citep{Kingma13} the negative log likelihood (NLL) term in the objective punishes reconstruction error forcing the model to learn relevant information on the belief state to be captured in the posterior output, while the KL term pulls the prior output towards the posterior output, forcing the prior and posterior to agree on the content of the belief state in a way that still allows the likelihood model to reconstruct the current observation.

We can now use the learned models to engage in active inference, and infer which action the agent has to take next. This is done by generating imagined trajectories for different policies using $p_\theta$ and $p_\xi$, calculating the expected free energy $G$ and selecting the action of the policy that yields the lowest $G$. These policies to evaluate can be predefined, or generated through random shooting, using cross-entropy method~\citep{BoerCEM} or by building a search tree.

\section{Experiments}\label{results}
We validate our deep active inference approach on a real world robotics navigation task. First, we collect a dataset consisting of two hours worth of real world action-observation sequences by driving a Kuka Youbot base platform up and down the aisles of a warehouse lab. Camera observations are recorded with a front mounted Intel Realsense RGB-D camera, without taking into account the depth information. The x, y and angular velocities are recorded as actions at a recording frequency of 10Hz. The models are trained on a subsampled version of the data resulting in a train set with data points every 200ms.

Next, we instantiate neural networks $q_\phi$ and $p_\xi$ as a convolutional encoder and decoder network, and $p_\theta$ using an LSTM. These are trained with Adam optimizer using the objective function from Equation~\ref{eq:model-obj} for 1M iterations. We use a minibatch size of 128 and a sequence length of 10 timesteps. A detailed overview of all hyperparameters is given in appendix.

We utilise the same approach as in~\cite{catal2020learning} for our imaginary trajectories and planning. The agent has access to three base policies to pick from: drive straight, turn left and turn right. Actions from these policies are propagated to the learned models at different time horizons $H = 10, 25$ or $55$. For each resulting imaginary trajectory, the expected free energy $G$ is calculated. Finally the trajectory with lowest $G$ is picked, and the first action of the chosen policy is executed, after which the imaginary planning restarts. The robot's preferences are given by demonstration, using the state distribution of the robot while driving in the middle of the aisle. This should encourage the robot to navigate in the aisles.

\begin{figure}[t!]
    \centering
    \begin{subfigure}[t]{0.45\textwidth}
        \includegraphics[width=0.9\textwidth]{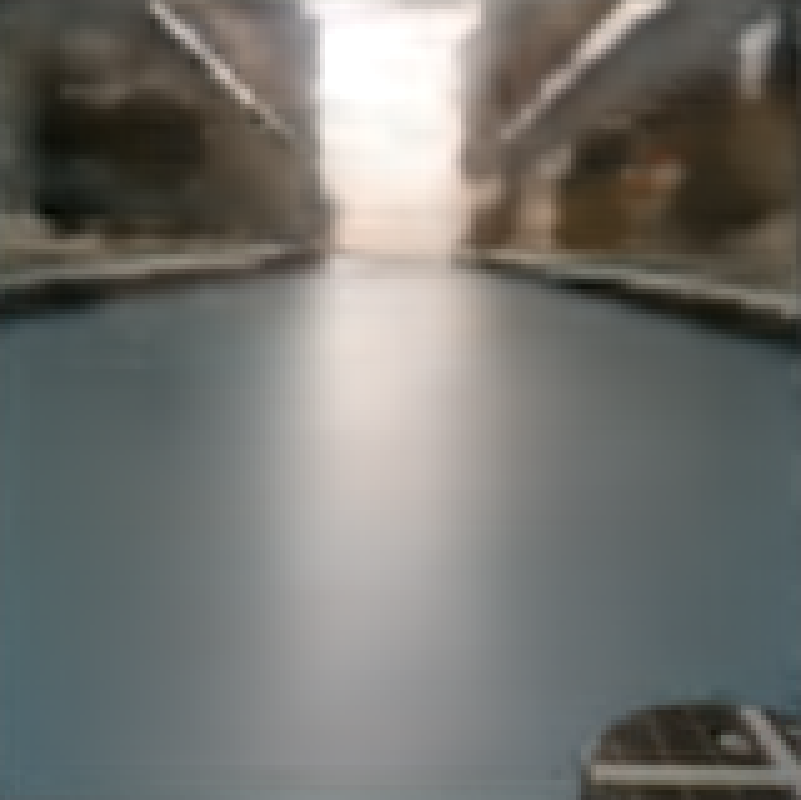}
        \caption{Preferred state.}\label{fig:preferred}
    \end{subfigure}\quad
    \begin{subfigure}[t]{0.45\textwidth}
        \includegraphics[width=0.9\textwidth]{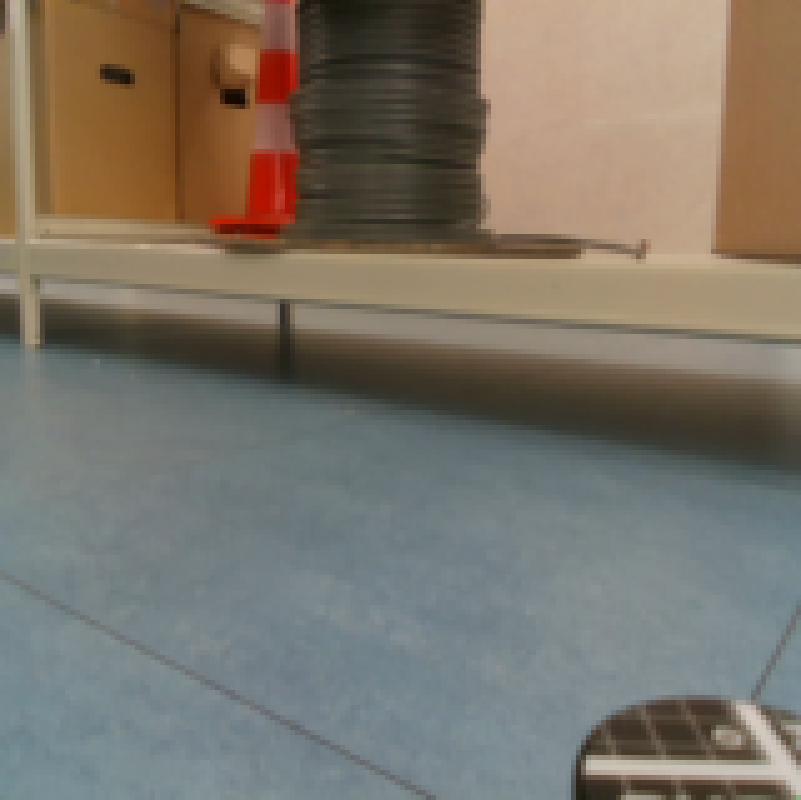}
        \caption{Start state}\label{fig:start_state}
    \end{subfigure}
    \begin{subfigure}[t]{\textwidth}
        \includegraphics[width=\textwidth]{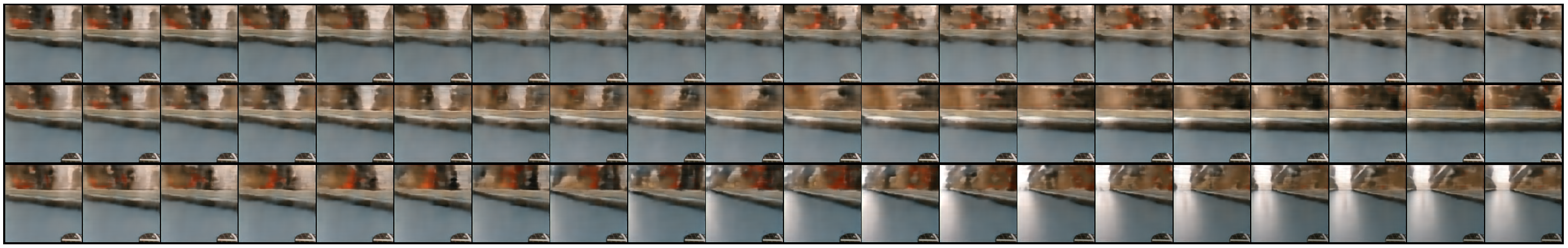}
        \caption{Imaginary future trajectories for different policies, i.e. going straight ahead (top), turning right (middle), turning left (bottom).}\label{fig:g-rollouts}
    \end{subfigure}\quad
    \begin{subfigure}[t]{\textwidth}
        \includegraphics[width=\textwidth]{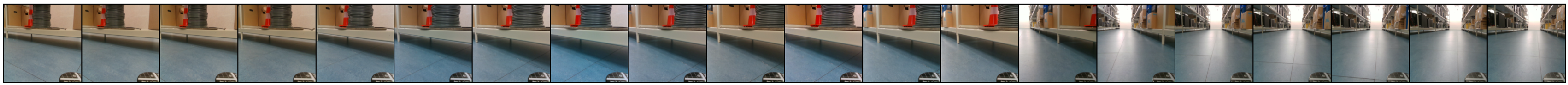}
        \caption{Actually followed trajectory.}\label{fig:trajectory}
    \end{subfigure}
    \caption{Experimental results: Figure (a) shows the target observation in imagined (reconstructed) space. (b) The start observation of the trial. Figure (c) shows different imaginary planning results, whilst (d) shows the actually followed trajectory.}\label{fig:policy-learning}
\end{figure}

At each trial the robot is placed at a random starting position and random orientation and tasked to navigate to the preferred position. Figure~\ref{fig:policy-learning} presents a single experiment as an illustrative example. Figure~\ref{fig:preferred} shows the reconstructed preferred observation from the given preferred state, while Figure~\ref{fig:start_state} shows the trial's start state from an actual observation. Figure~\ref{fig:g-rollouts} shows the imagined results of either following the policy ``always turn right'', ``always go straight'' or ``always turn left''. Figure~\ref{fig:trajectory} is the result of utilising the planning method explained above. Additional examples can be found in the supplementary material.

The robot indeed turns and keeps driving in the middle of the aisle, until it reaches the end and then turns around~\footnote{A movie demonstrating the results is available at  \url{https://tinyurl.com/smvyk53}}. When one perturbs the robot by pushing it, it will again recover and continue to the middle of the aisle.

\section{Conclusion}\label{discussion}
In this paper we present how we can implement a generative model for active inference using deep neural networks. We show that we are able to successfully execute a simple navigation task on a real world robot with our approach. As future work we want to allow the robot to continuously learn from past autonomous behaviour, effectively ``filling the gaps'' in its generative model. Also how to define the ``preferred state'' distributions and which policies to evaluate remains an open research challenge for more complex tasks and environments. 



\bibliography{main}
\bibliographystyle{iclr2020_conference}
\newpage
\appendix
\appendixpage

\section{Neural architecture}

\begin{table}[h!]
    \centering
    \begin{tabular}{l | c | c | c  }
        & Layer & Neurons/Filters & activation function  \\\hline \hline
        \parbox[t]{2mm}{\multirow{7}{*}{\rotatebox[origin=c]{90}{Posterior}}}
        &Convolutional & 8   & Leaky ReLU \\
        &Convolutional & 16  & Leaky ReLU \\
        &Convolutional & 32  & Leaky ReLU \\
        &Convolutional & 64  & Leaky ReLU \\
        &Convolutional & 128 & Leaky ReLU \\
        &Concat        & N.A. & N.A.      \\
        &Linear        & 2 x 128 states & Softplus\\ \hline

        \parbox[t]{2mm}{\multirow{6}{*}{\rotatebox[origin=c]{90}{Likelihood}}}
        & Linear &  128 x 8 x 8  & Leaky ReLU \\
        &Convolutional & 128  & Leaky ReLU  \\
        &Convolutional & 64   & Leaky ReLU  \\
        &Convolutional & 32   & Leaky ReLU  \\
        &Convolutional & 16   & Leaky ReLU  \\
        &Convolutional & 8    & LeakyReLU   \\ \hline

         \parbox[t]{2mm}{\multirow{3}{*}{\rotatebox[origin=c]{90}{Prior}}}
        &LSTM cell &  400  & Leaky ReLU \\
        &Linear & 2 x128 states  & Softplus \\
        &       &                                  &          \\
    \end{tabular}
    \caption{Neural network architectures. All convolutional layers have a 3x3 kernel. The convolutional layers in the Likelihood model have a stride and padding of 1 to ensure that they preserve the input shape. Upsampling is done by nearest neighbour interpolation. The concat step concatenates the processed image pipeline with the vector inputs $\va$ and $\vs$.}
    \label{tab:Architectures}
\end{table}

\newpage
\section{Hyperparameters}
\begin{table}[h!]
    \centering
    \begin{tabular}{l|c|c}
        &Parameter  & Value \\ \hline \hline
        \parbox[t]{2mm}{\multirow{4}{*}{\rotatebox[origin=c]{90}{Learning}}}
        &learning rate & 0.0001 \\
        &batch size & 128 \\
        &train iterations & 1M \\
        &sequence length & 10 \\ \hline
        \parbox[t]{2mm}{\multirow{5}{*}{\rotatebox[origin=c]{90}{Planning}}}
        &$\gamma$ & 100 \\
        &D~\citep{catal2020learning} & 1 \\
        &K~\citep{catal2020learning} & 10, 25, 55 \\
        &N~\citep{catal2020learning} & 5 \\
        &$\rho$~\citep{catal2020learning} & 0.001

    \end{tabular}
    \caption{Overview of the model hyperparemeters.}
    \label{tab:my_label}
\end{table}


\newpage
\section{Detailed Planning example}
A movie demonstrating the results is available at \url{https://tinyurl.com/smvyk53}.
\begin{figure}[h!]
    \centering
    \includegraphics[width=\textwidth]{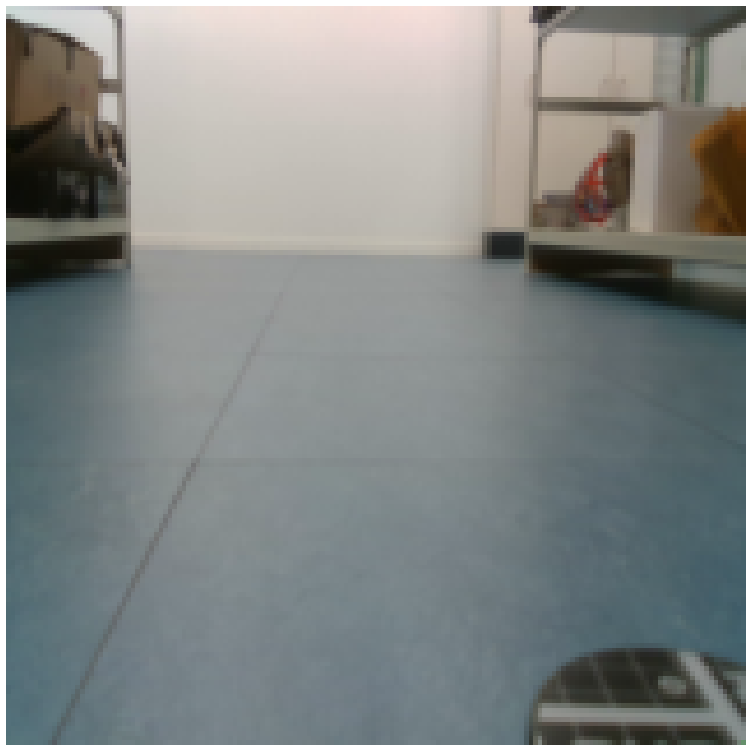}
    \caption{Trial preferred state}
    \label{fig:large-pref}
\end{figure}

\begin{figure}
    \centering
    \includegraphics[width=\textwidth]{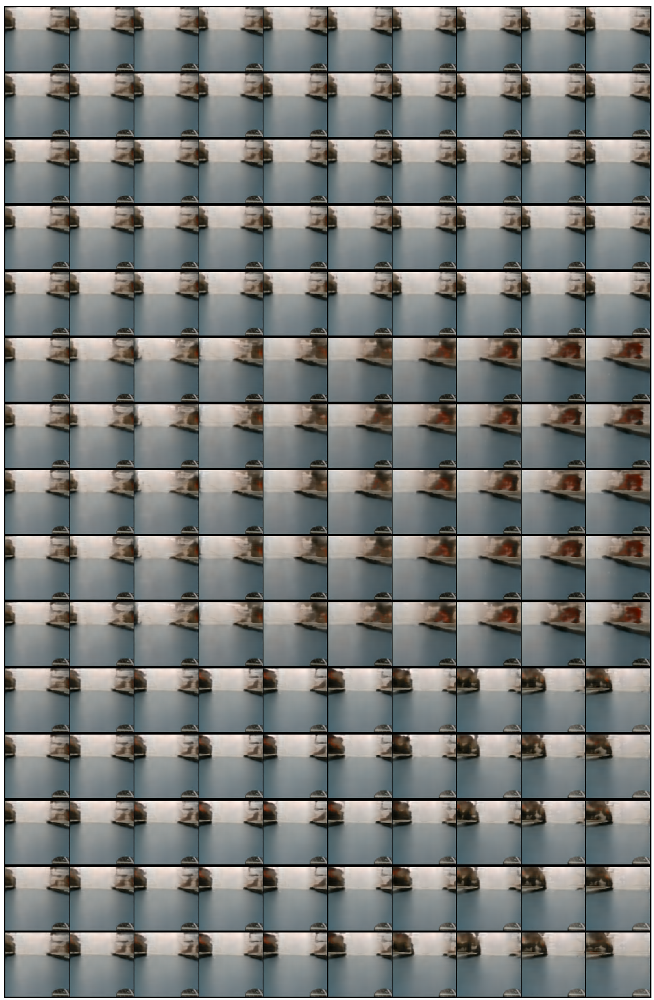}
    \caption{Short term planning}
    \label{fig:large-short-term}
\end{figure}

\begin{sidewaysfigure}
    \centering
    \includegraphics[width=\linewidth]{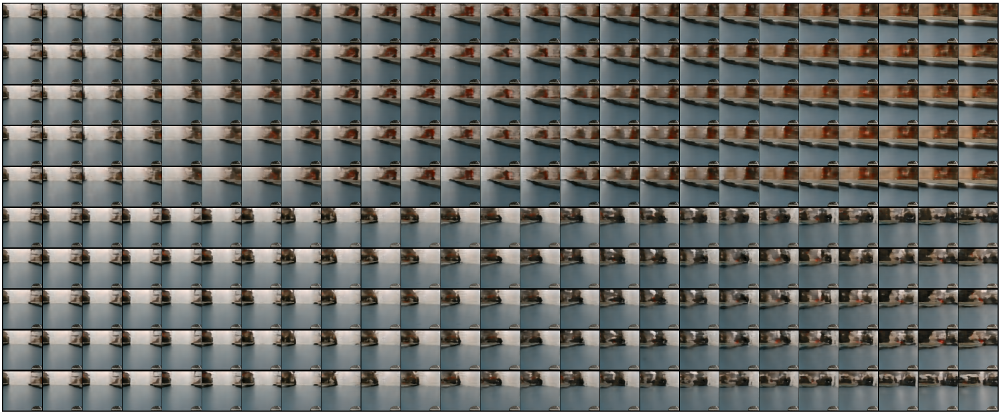}
    \caption{Middle long term planning}
    \label{fig:large-mid-term}
\end{sidewaysfigure}

\begin{sidewaysfigure}
    \centering
    \includegraphics[width=\linewidth]{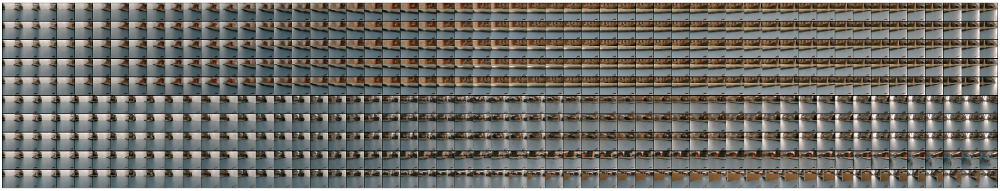}
    \caption{Long term planning}
    \label{fig:large-long-term}
\end{sidewaysfigure}

\end{document}